\documentclass[10pt,twocolumn,letterpaper]{article}

\usepackage[pagenumbers]{cvprAuthorKit/cvpr} 

\usepackage{graphicx}
\usepackage{amsmath}
\usepackage{amssymb}
\usepackage{booktabs}

\usepackage[pagebackref,breaklinks,colorlinks]{hyperref}

\def\mypar#1{\vspace{1mm}{\noindent\bf #1.}\hspace{1mm}}

\def\fig#1{Fig.~\ref{fig:#1}}
\def\myvspace{\vspace{1mm}} 

\def\cvprPaperID{***} 
\def\confName{CVPR}
\def\confYear{2023}

\begin{document}


\title{Are Visual Recognition Models Robust to Image Compression?}

\author{
João Maria Janeiro$^{1}$  \quad
Stanislav Frolov$^{2,4}$  \quad
Alaaeldin El-Nouby$^{1,3}$
Jakob Verbeek$^{1}$
\\
$^1$Meta AI\quad
$^2$RPTU Kaiserslautern-Landau\quad
$^3$Inria\quad\\
$^4$German Research Center for Artificial Intelligence (DFKI)
}

\maketitle
\begin{abstract}
Reducing the data footprint of visual content via image compression is essential to reduce storage requirements, but also to reduce the bandwidth and latency requirements for transmission.
In particular, the use of compressed images allows for faster transfer of data, and faster response times for visual recognition in edge devices that rely on cloud-based services.
In this paper, we first analyze the impact of  image compression using traditional codecs, as well as recent state-of-the-art neural compression approaches, on three visual recognition tasks: image classification, object detection, and semantic segmentation.
We consider a wide range of compression levels, ranging from 0.1 to 2 bits-per-pixel (bpp). 
We find that for all three tasks, the recognition ability is significantly impacted when using strong compression.
For example, for segmentation mIoU is reduced from  44.5 to 30.5 mIoU when compressing to 0.1 bpp using the best compression model we evaluated.
Second, we test to what extent this performance drop can be ascribed  to a loss of relevant information in the compressed image, or to a lack of generalization of visual recognition models to images  with compression artefacts. 
We find that to a large extent the performance loss is due to the latter: by finetuning the recognition models on compressed training images, most of the performance loss is recovered. 
For example, bringing segmentation accuracy back up to 42 mIoU, i.e.  recovering 82\% of the original drop in accuracy.
\end{abstract}

\section{Introduction}

Mobile devices with high resolution vision sensors, but limited  storage and compute  capabilities, are ubiquitous:  including smartphones, watches, and AR/VR devices.
Image compression is  critical  to facilitate storage of the captured data on-device, and to reduce  the required channel bandwidth and latency for remote storage.
State-of-the-art  recognition models that enable analysis of visual  data, rather than just storing it, are currently without exception based on deep learning.
They  impose heavy memory and compute requirements, despite significant efforts to reduce inference cost, \eg using efficient architectures~\cite{howard2017mobilenets, iandola2016squeezenet, tan19icml}, 
weight compression~\cite{masana2017domain,tai2015convolutional},  quantization~\cite{fan21iclr,jacob2018quantization,lin2016fixed}, and network pruning~\cite{ghosh2018structured,lecun90nips,li2016pruning,veniat18cvpr}.
The use of state-of-the-art vision models for low-latency applications, therefore, requires  transmission of the data to compute servers in compressed format, and recognition models should be robust to artefacts that may be introduced by compression.

\begin{figure}
    \begin{center}
       \includegraphics[width=\linewidth]{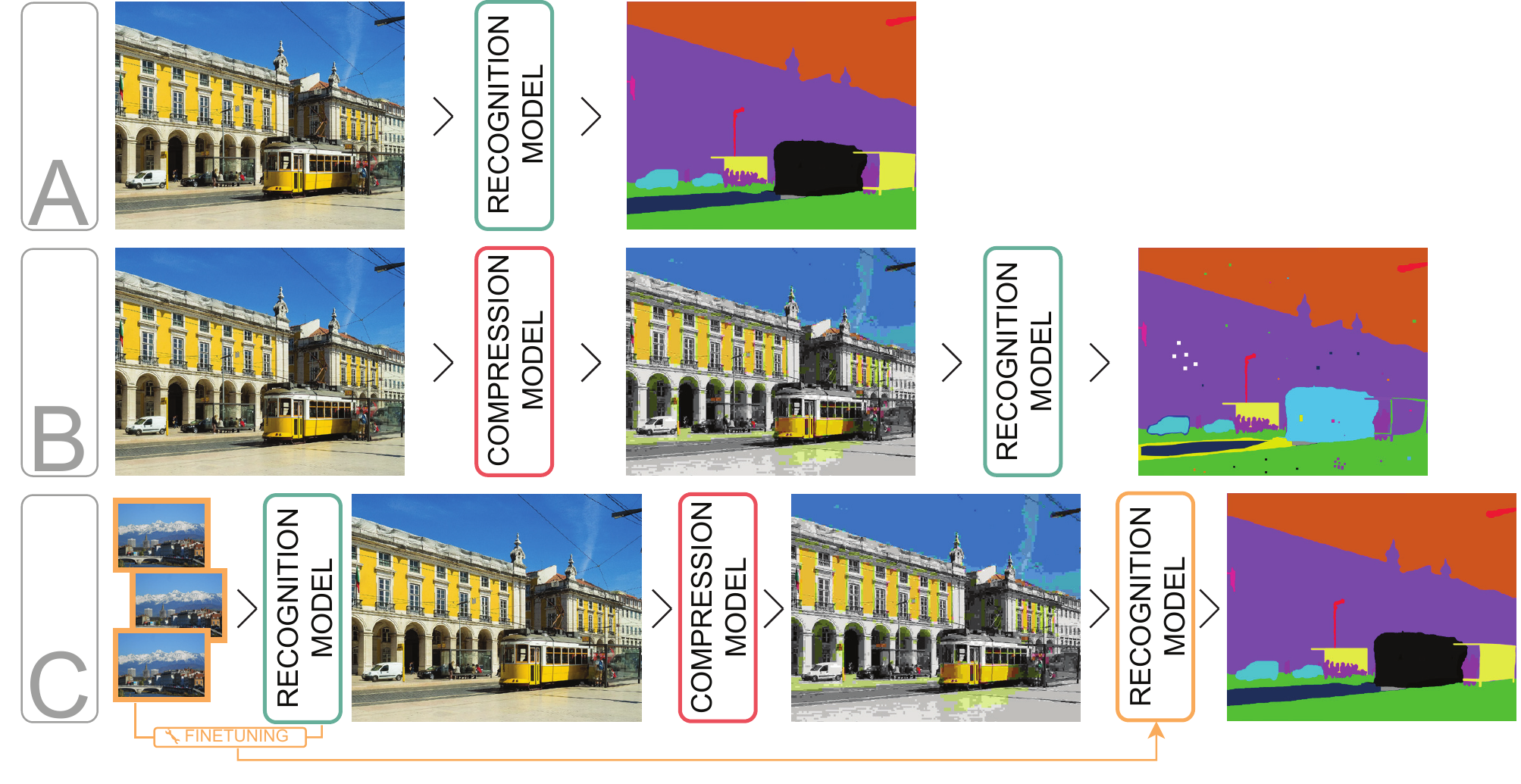}
    \end{center}
    \caption{Illustration of the three scenarios we consider. 
    In  A, the recognition model is trained and evaluated on the original dataset  images: this is our reference baseline. 
    In B, we use the model from A,  but evaluate its performance on  compressed images. 
    In C, the recognition model is finetuned on compressed images, and then tested the same way as option B.
    }
    \label{fig:teaser}
\end{figure}

Prior works have focused on faster and more efficient processing, by learning vision recognition decoders directly on compressed features \cite{park22arxiv, wiles22accv}.
Another focus has been on faster transfer of data, through split computation with compressed data \cite{choi2018, nakahara2021}.
The aforementioned works require architectural changes to the networks, and novel methods.
Moreover, previous work mostly considers  a single compression algorithm, JPEG in \cite{park22arxiv}, HEVC \cite{Sullivan12HEVC} in \cite{choi2018}, and VQ-VAE in \cite{wiles22accv}.
To the best of our knowledge,  the impact of image compression  on visual recognition has not been systematically studied.

In this work, we evaluate to what extent state-of-the-art visual recognition models are robust to compression of the input images across three tasks: image classification, object detection and semantic segmentation, on ImageNet \cite{deng09cvpr}, COCO \cite{caesar18cvpr} and ADE20K \cite{zhou17cvpr}, respectively.
We explore both neural compression methods, as well as traditional hand-engineered codecs.  
We  consider bitrates from 2 bits-per-pixel (bpp) down to 0.1 bpp, ranging from  high-quality compression to an extreme compression regime where visible  artefacts are introduced.

We find that for all tested  codecs, image compression leads to a degradation of visual recognition performance, in particular at low bitrates.
A-priori, it is not clear what is causing the degradation: compression can lead to a loss of detail which makes the recognition tasks intrinsically harder, or the recognition models  do not generalize well to compressed images due to a lack of robustness to the domain shift introduced by the compression artefacts.
By finetuning the recognition models on compressed images we can mitigate the domain shift, and test what causes the observed performance degradation.
We find that most of the performance loss can be recovered using the finetuned models, suggesting that the  performance reduction can be attributed to the models' inability to generalize to images with compression artefacts, rather than the presence of compression artefacts increasing the difficulty of recognition. 
The experimental setup that we used to conduct our evaluations is illustrated in \fig{teaser}.

To summarize, we make the following contributions:
\begin{itemize}
  \setlength\itemsep{0mm}
    \item We evaluate the impact on image classification, object detection and semantic segmentation accuracy, when compressing images with state-of-the-art traditional as well as learned neural codecs.
    \item We observe significant degradations in recognition accuracy in the very low bitrate regime of 0.1 bpp, and find that this is mostly caused by the inability of recognition models to generalize to images with compression artefacts.
    \item We show that most of the accuracy loss can be recovered by finetuning recognition models on compressed images, in particular when using neural compression.   
    For detection and segmentation with finetuning, the mAP and mIoU obtained using original images can be approximated up to 0.5 points with images compressed to 0.4 bpp, reducing the image data size by a factor 4 and 12 for  segmentation and detection, respectively.
\end{itemize}

\begin{figure*}
    \centering
    \includegraphics[width=.3\textwidth]{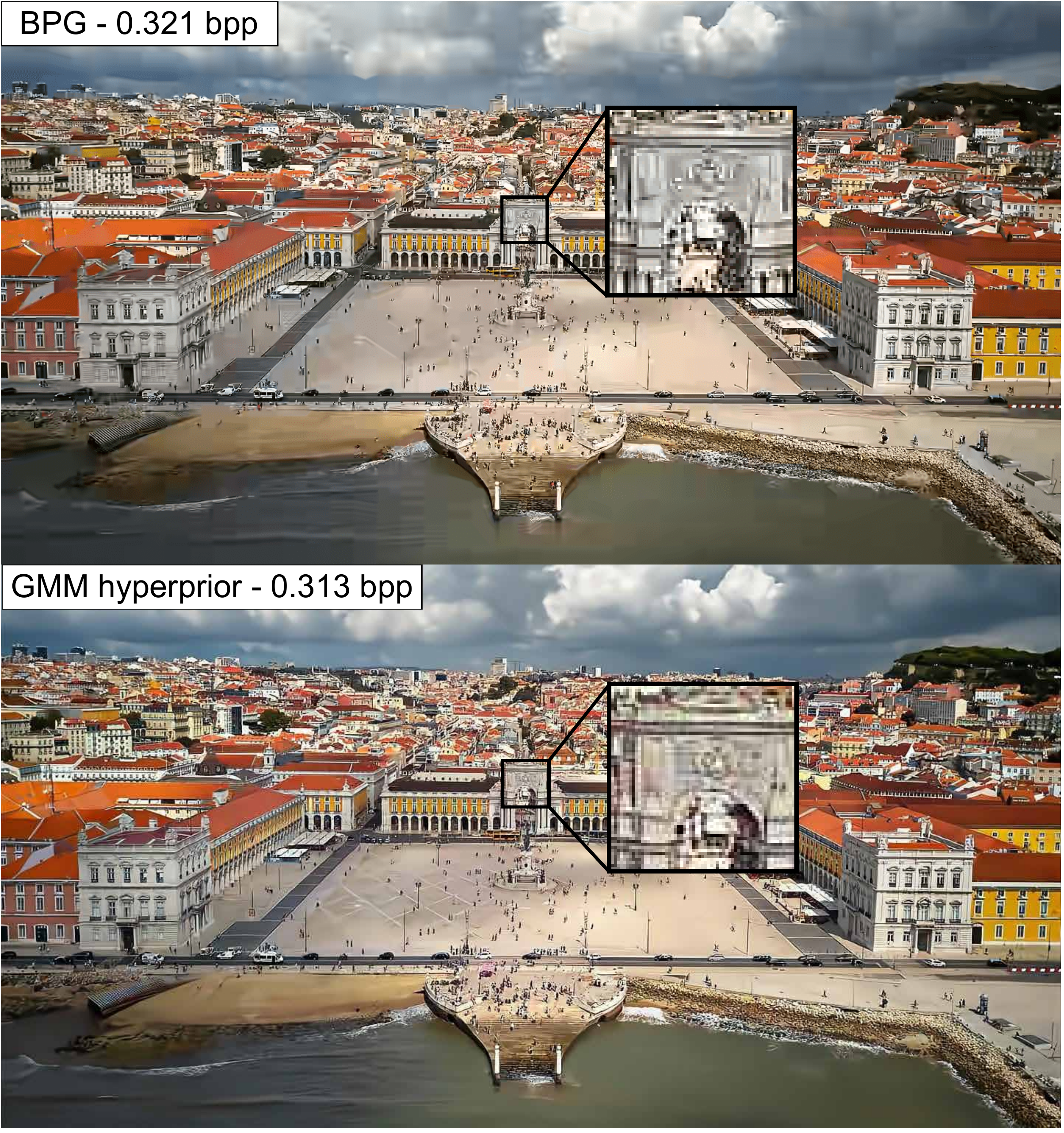}\hfill
    \includegraphics[width=.3\textwidth]{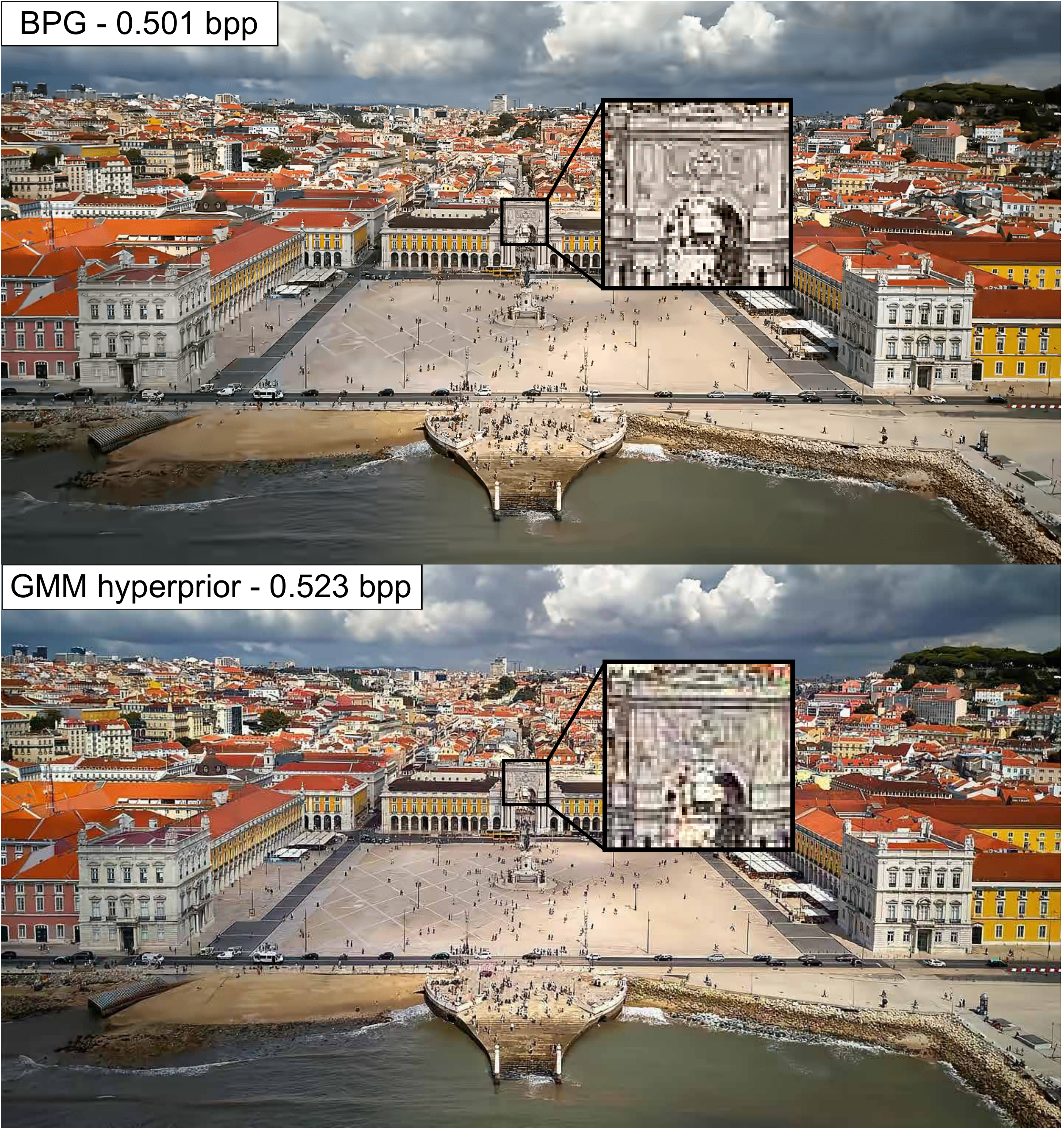}\hfill
    \includegraphics[width=.3\textwidth]{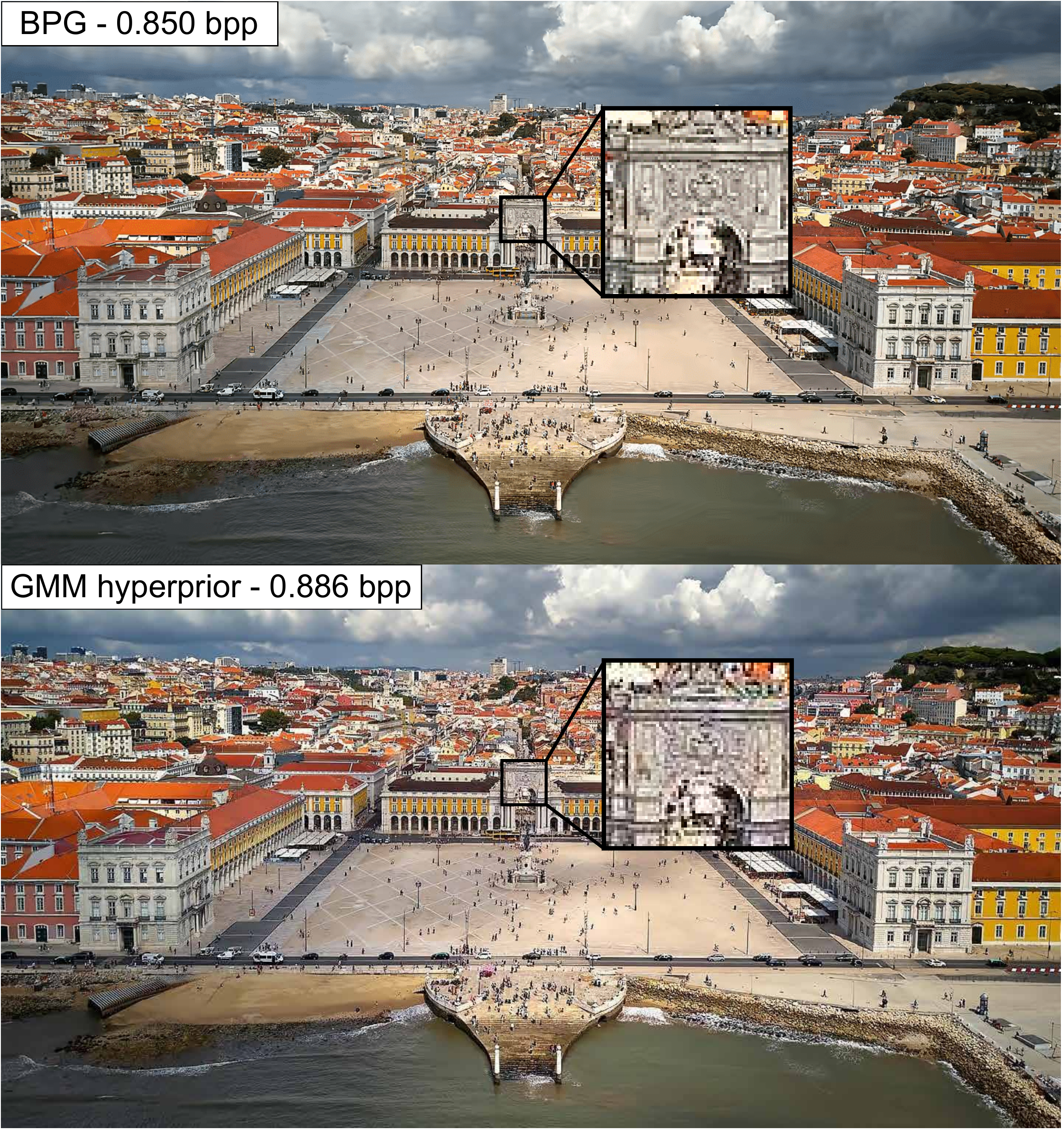}
    \myvspace
    \caption{Image compressed at three different bitrates using  BPG and GMM hyperprior. 
    Black  suqare provides a zoom  of the central area.
    }
    \label{fig:compression}
\end{figure*}

\section{Related work}

\mypar{Neural compression methods}
Most neural image compression methods follow an autoencoder architecture as a way to achieve a good reconstruction from a small latent representation  space, see 
\eg~\cite{balle18iclr, wiles22accv, el2022image,mentzer19cvpr,rippel17icml}.
An entropy model is employed 
to estimate the probability distribution of the quantized latent representation,
which is in turn used by an entropy coder ---typically an arithmetic coder--- to compress the latent representation in a lossless manner into a bit stream, see \eg~\cite{mackay03book}.

\mypar{Vision tasks from compressed latent space}
Several prior works have explored learning visual recognition models from  compressed latent representations~\cite{park22arxiv, wiles22accv, 10.1145/3559105}.
For example,~\cite{park22arxiv} trains a ViT~\cite{dosovitskiy21iclr} directly on JPEG coefficients, and expresses common data augmentations in the same space.
They evaluate on ImageNet classification~\cite{deng09cvpr}, and   achieve similar performance to the RGB model.
On the other hand,~\cite{10.1145/3559105} instead trains a CNN on the frequency-domain features, and assesses its performance in object detection and image classification tasks.
For video,~\cite{wiles22accv} uses a VQ-VAE autoencoder~\cite{oord17nips,razavi19nips} at the frame level, and learns video classification models  on the bottleneck representation. This reduces memory and compute requirements, allowing processing of minute to hour long videos.

\mypar{Split computing with compression}
The computation of a model can be divided between the user's device and the cloud.
Several works make use of image/feature compression for faster data transfers~\cite{choi2018, nakahara2021, cohen2021lightweight}.
In~\cite{choi2018} an object detection model is  trained to compensate for the lossy feature compression artefacts.
Transmission of an image compressed at different bitrates until the desired recognition quality is achieved is explored in~\cite{nakahara2021}.

\vspace{1mm}
All the aforementioned works focus on a single compression method, and develop new techniques for a single  task.
Meanwhile, our focus is not to create a new method, but rather to systematically evaluate existing compression methods for several representative recognition tasks.

\section{Experimental setup}
This section covers the compression methods employed in this study, the recognition tasks used to evaluate their effectiveness, as well as their training and testing setup.

\subsection{Image compression codecs}

 We use four state-of-the-art compression codecs: two traditional compression codecs, BPG~\cite{ballard_bpg} and WebP~\cite{webp}, and two neural compression methods based on the hyperprior model~\cite{balle18iclr}. 
In particular, we use the Mean and Scale (M\&S) hyperprior model~\cite{minnen18nips}, 
and the Gaussian Mixture Model (GMM) hyperprior \cite{cheng20cvpr}.
M\&S combined a mean and scale hyperprior with an autoregressive context model, for better rate-distortion trade-offs.
GMM improves over M\&S by replacing the  Gaussian  likelihood model over the latents by a Gaussian mixture model, which better captures the conditional distributions given the hyperlatents.
In \fig{compression}, we present an image compressed at three different rates by  BPG and GMM hyperprior, to illustrate the image quality and artefacts at the  bitrates conside
0red in our experiments.
We utilize the PIL library~\cite{pil} for WebP, Bellard's implementation for BPG~\cite{ballard_bpg}, and the CompressAI library~\cite{begaint2020compressai}  library for the neural codecs.
To compute the image sizes in bit-per-pixel (bpp), we use the CompressAI library for the neural codecs and WebP, while  for BPG we divide the image file size by the number of pixels. 
For the  original images in the datasets, we compute the bpp based on the  JPEG  filesizes. 

\subsection{Visual recognition tasks}

We consider image classification, object detection and semantic segmentation as representative  recognition tasks.
For classification and segmentation, we use a Swin-T  backbone~\cite{Liu_2021_ICCV}, combined with an MLP head for classification, and an UPerNet head~\cite{xiao18eccv} for segmentation.
For detection, the backbone is a ResNet-50 \cite{he16cvpr}, with a Disentangled Dense Object Detector (DDOD) head~\cite{chen2021disentangle}.
We use implementations of the MMClassification~\cite{2020mmclassification},   MMDetection~\cite{mmdetection}, and MMSegmentation~\cite{mmseg2020} libraries. 
We evaluate the models on  ImageNet~\cite{deng09cvpr} for classification, COCO~\cite{caesar18cvpr} for detection, and ADE20K~\cite{zhou17cvpr} for segmentation.
For each task we use the standard evaluation metrics: 
 accuracy for classification, mean average precision (mAP) for detection, and  mean intersection-over-union (mIoU) for segmentation.

In our experiments we evaluate the public checkpoints released for the different models in the corresponding libraries, which are trained on the original images in the datasets.
We experimentally observe that the recognition accuracy of these models  deteriorates when evaluated on compressed images. This could be due to a loss of detail when compressing, which makes the recognition tasks intrinsically harder, or because the recognition models lack robustness and do not generalize well to compressed images.
To investigate how these factors contribute,  we finetune the models using compressed versions of the training images, so that the models adapt to compression artefacts, and the original domain shift in the input data is eliminated. 

In practice, we use the same amount of finetuning iterations as were originally used to adapt the pre-trained backbones to the different tasks. 
For classification the model finetuned for 30 epochs, for detection 12 epochs, and for segmentation 160k iterations. 
We finetune models separately for each compression level.

To factor out the influence of additional training, we also finetune the baseline models on the original datasets, for the same amount of additional epochs.
We select the best scoring model, original or finetuned, as the baseline.
For  classification and segmentation, finetuning the original model did not improve accuracy, while for detection finetuning did improve the original model.
\section{Experimental results}

\begin{figure}
    \begin{subfigure}{\linewidth}
          \includegraphics[width=\linewidth]{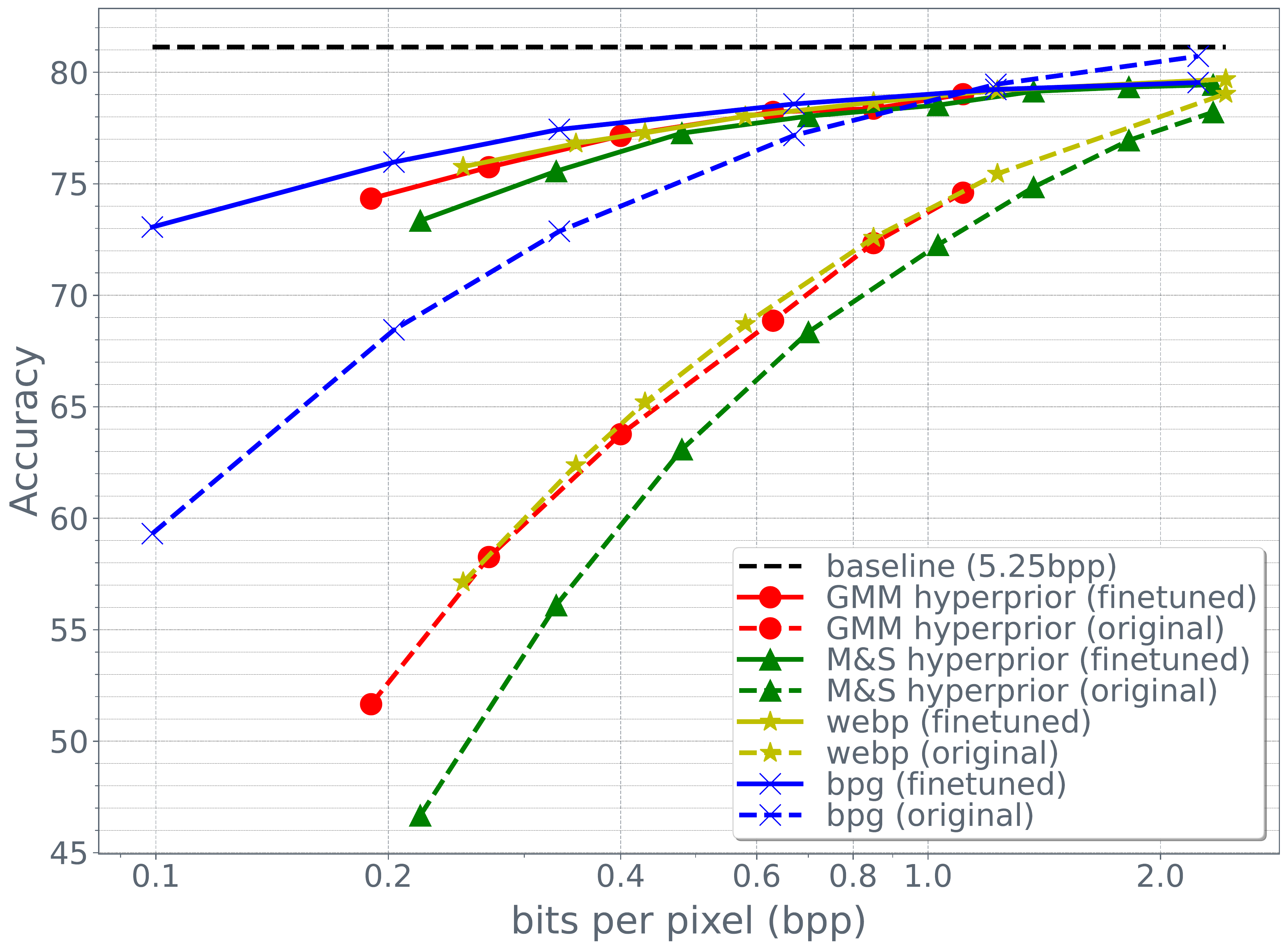}
          \caption{Results for image classification on ImageNet.}
          \label{fig:results/classification}
    \end{subfigure}%
\\
    \begin{subfigure}{\linewidth}
          \includegraphics[width=\linewidth]{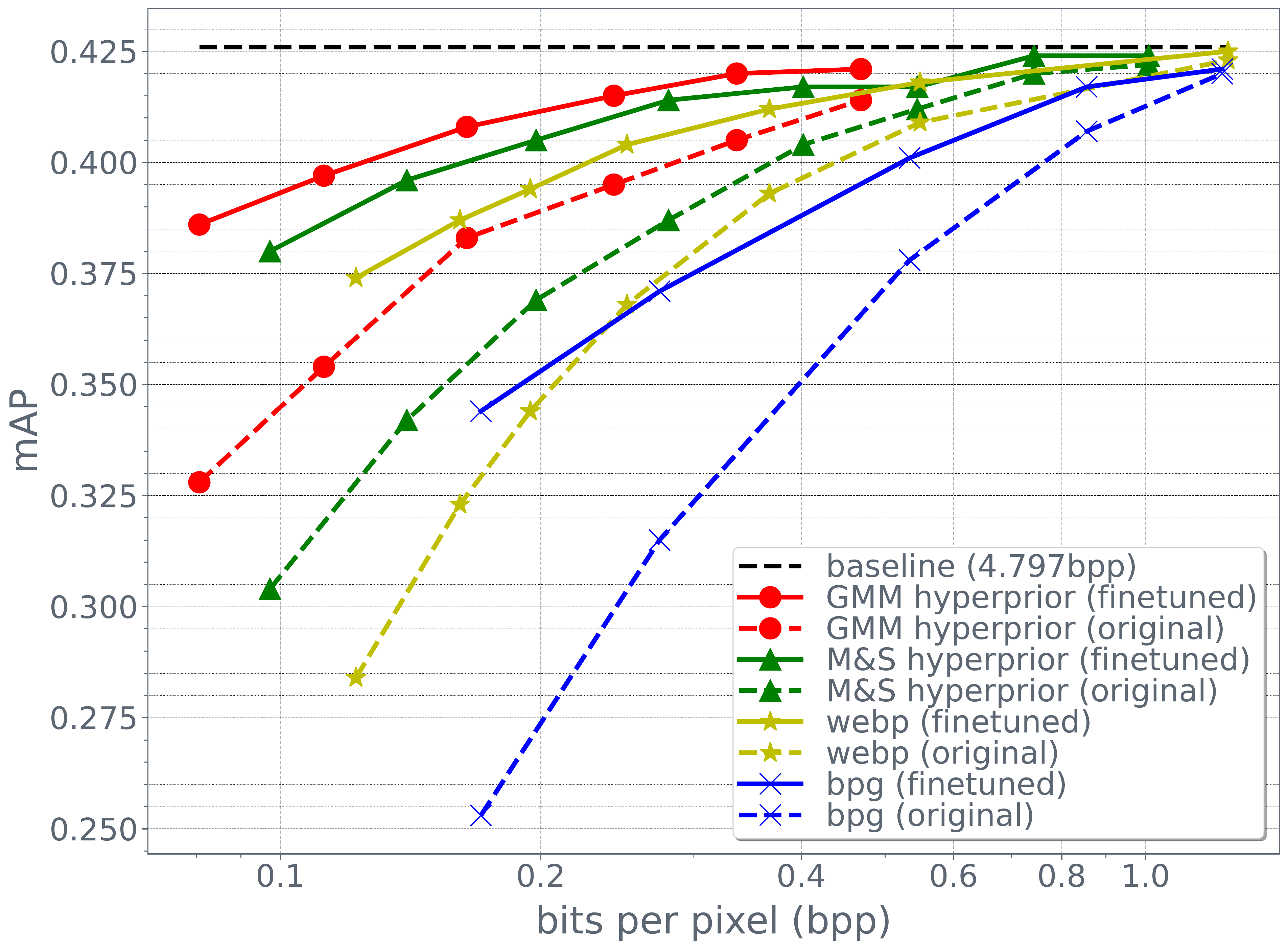}
          \caption{Results for object detection on COCO.}
          \label{fig:results/detection}
    \end{subfigure}
    \\
    \begin{subfigure}{\linewidth}
          \includegraphics[width=\linewidth]{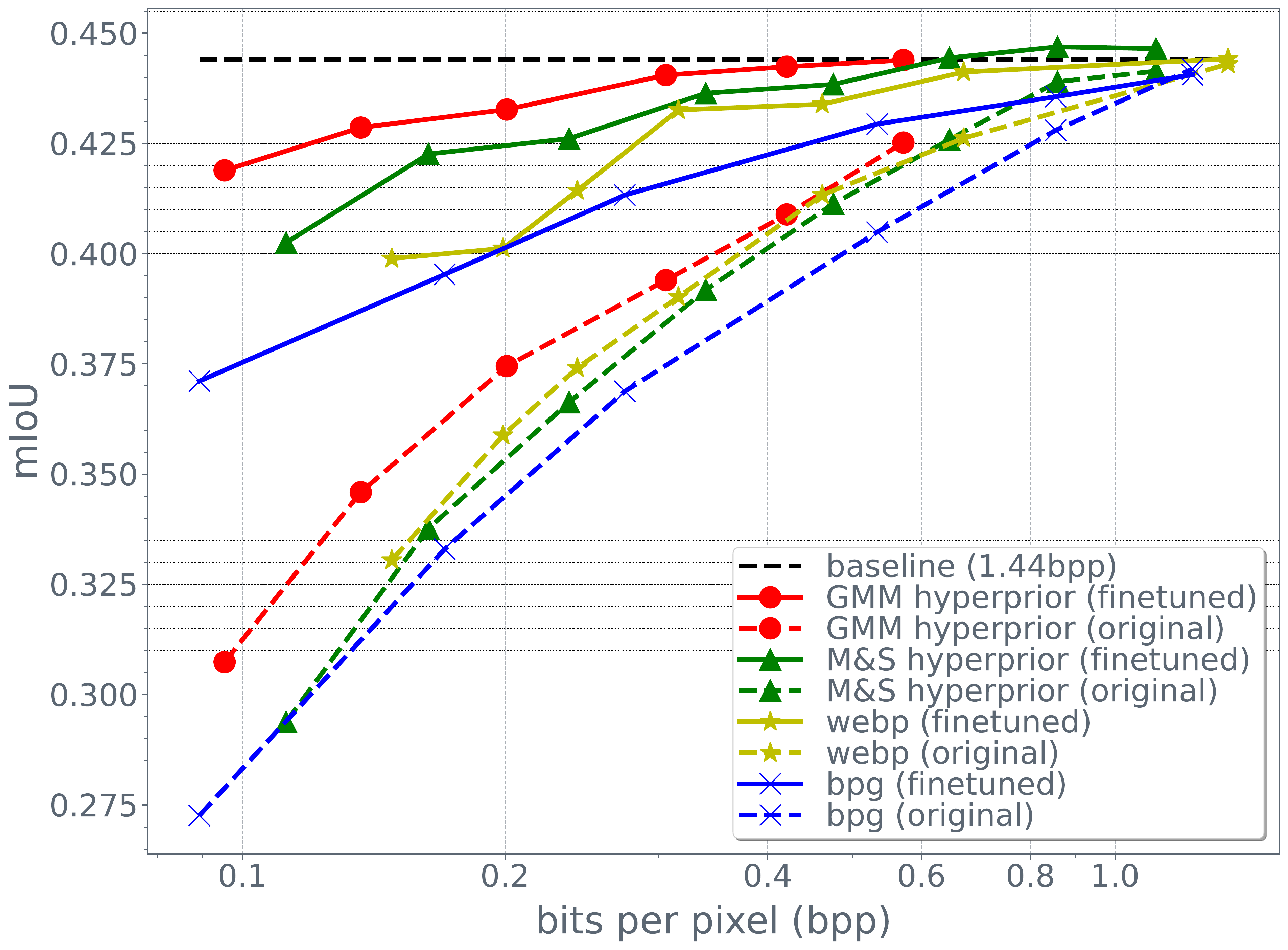}
          \caption{Results for semantic segmentation on ADE20K.}
          \label{fig:results/segmentation}
    \end{subfigure}
    \caption{Visual recognition results with compressed images. 
    The horizontal dashed black line is the baseline result obtained using the original images.
    Other curves evaluate  models trained on original images (original, dashed), and model finetuned using compressed images (finetuned, solid), test images are compressed using WebP, BPG, M\&S and GMM hyperprior codecs. 
    }
      \label{fig:results_plots}
\end{figure}

We present our main experimental results in \fig{results_plots}, and discuss and interpret the results below.

\mypar{Classification} 
When using the baseline model trained on the original images for classification (dashed curves in \fig{results/classification}),  we found that compressing images with BPG has the least impact on recognition accuracy, followed by WebP and GMM hyperprior which yield comparable impacts. 
Finetuning the  model on compressed images (solid curves) yields a significant improvement in results. 
For example, improving accuracy from 59.5\% to 73\% for BPG compression at  0.1 bpp, relative to baseline accuracy of 81\%  on the original images (5.2 bpp).
This shows that, to a large extent, the accuracy drop observed when testing on compressed images, is due to the lack of generalization of the original model to images with compression artefacts.
After finetuning, at 1 bpp the accuracy is around 79\% for all compression methods; a 3\% loss \wrt the baseline model while reducing the bitrate by a factor five.

\mypar{Object detection} 
Interestingly, the results for object detection on COCO in  \fig{results/detection}, show a different ordering of results \wrt the different compression codecs. 
Here, the traditional codecs BPG and WebP lead to bigger drops  in accuracy  than the neural compression models.
In fact, even when finetuning on compressed images the results for BPG (solid blue) are  worse than using the original model on  images compressed using the neural codecs (dashed green and red). 
Similar to the classification experiment, the drop in object detection accuracy can to a large extent be recovered by finetuning the model on compressed images. 
For example, for the neural codecs at 0.1 bpp, the initial drop of 10 points or more in mAP is reduced to under 5 points.
At 0.4 bpp, after finetuning the GMM hyperprior model is able to reduce the bit rate by more than a factor 10, while reducing the mAP by only 0.5 (from 42.5 to 42.0) \wrt the baseline model on the original images.

\mypar{Semantic segmentation} 
For semantic segmentation we observe  similar trends as for detection:  BPG compression hurts accuracy most, and  GMM hyperprior compression has least impact. 
When compressing images with the GMM hyperprior codec to 0.1 bpp, an mIoU of 31\% is obtained using the baseline model, while the finetuned model obtains 42\%. 
In comparison, the baseline model on the original images (1.44 bpp) obtains 44.5\%.
At 0.6 bpp the mIoU of the finetuned model on GMM hyperprior compressed images matches the performance of the baseline model on the original images. 
\section{Conclusion}
We investigated the impact of image compression on visual recognition, using both traditional codecs and  recent neural compression methods  for compression levels ranging from moderate (2 bpp) to very strong compression (0.1 bpp). 
We find that strong compression has a big negative impact on the accuracy for tasks such as image classification, object detection and semantic segmentation. 
Our experiments show that this is to a large extent due to the lack of generalization of these models to images with compression artefacts. 
By finetuning the recognition models on compressed images, we find that most of the loss in accuracy on compressed images can be recovered.

Our findings can contribute to deploy visual recognition for users in resource and bandwidth limited settings. 
In future work we want to explore to what extent our findings can be used to reduce I/O bound latency when training visual recognition models on internet-scale datasets. 
In particular, it is interesting to explore training recognition models directly on the latent compressed image representations, rather than passing through the usual RGB representation.

\medskip
\mypar{Photo credits}
Figure 1 main photo by Ajay Suresh, under CC License 2.0 \cite{cc2_license}.
Figure 1 small photo by Zhaoshan75, under CC License 4.0 \cite{cc4_license}.
Figure 2 by Deensel, under CC License 2.0 \cite{cc2_license}.

{\small
\bibliographystyle{cvprAuthorKit/ieee_fullname}
\bibliography{jjv}}

\end{document}